\documentclass[a4paper, 12pt]{article}
 
\usepackage[cp1252]{inputenc}
\usepackage[T1]{fontenc}
\usepackage{lmodern}
\usepackage{graphicx}
\usepackage[english, frenchb]{babel}
\usepackage[lined, ruled,linesnumbered]{algorithm2e}
\usepackage{graphicx}
\usepackage{subfigure}
\usepackage{epsfig}
\usepackage{amssymb}
\usepackage{amsmath}
\usepackage{amsfonts}
\usepackage{color}
\usepackage{authblk}

\begin{document}
%
%
\title{Dynamic recommender system: using cluster-based biases to improve the accuracy of the predictions}
\renewcommand\Affilfont{\itshape\small}
\setlength{\affilsep}{1em}
\author[1,2]{Modou Gueye}
\author[1]{Talel Abdessalem}
\author[3]{Hubert Naacke}
\affil[1]{Institut Telecom - Telecom ParisTech\authorcr 46, rue Barrault 75013\authorcr Paris, France\authorcr firstname.lastname@telecom-paristech.fr}
\affil[2]{Universit\'e Cheikh Anta DIOP\authorcr BP. 16432 Fann\authorcr Dakar, S\'en\'egal\authorcr gmodou@ucad.sn}
\affil[3]{LIP6, UPMC Sorbonne Universit\'es - Paris 6\authorcr 4, place Jussieu 75005\authorcr Paris, France\authorcr hubert.naacke@lip6.fr}
\maketitle
\selectlanguage{english}
\begin{abstract}
It is today accepted that matrix factorization models allow a high quality of rating prediction in recommender systems.
However, a major drawback of matrix factorization is its static nature that results in a progressive declining of the accuracy of the predictions after each factorization. This is due to the fact that the new obtained ratings are not taken into account until a new factorization is computed, which can not be done very often because of the high cost of matrix factorization.

In this paper, aiming at improving the accuracy of recommender systems, we propose a cluster-based matrix factorization technique that enables online integration of new ratings. Thus, we significantly enhance the obtained predictions between two matrix factorizations. We use finer-grained user biases by clustering similar items into groups, and allocating in these groups a bias to each user. The experiments we did on large datasets demonstrated the efficiency of our approach.
\end{abstract}
\section{Introduction}
The purpose of recommender systems is to predict user preferences on a large selection of items, i.e. find items that are likely to be of interest for the user. Because the user is often overwhelmed for facing the considerable amount of items provided by electronic retailers, the predictions are a salient function of all types of e-commerce~\cite{Schafer:1999:RSE:336992.337035,Bell:2009:MDP:1669704.1669716}. That is why recommender systems attract a lot of attention due to their great commercial value~\cite{Dias:2008:VPR:1454008.1454054,DBLP:conf/recsys/JannachH09,DBLP:journals/dsonline/LindenSY03,Fleder:2007:RSI:1250910.1250939}.

Collaborative filtering is a widely used category of recommender systems. It consists in analyzing relationships between users and interdependencies among items to identify new user-item associations~\cite{Su:2009:SCF:1592474.1722966,Koren:2009:MFT:1608565.1608614,citeulike:4563135}. Based on these associations, recommendations are inferred.
One of the most successful collaborative filtering algorithms is matrix factorization (MF). It gives good scalability and predictive accuracy~\cite{Takacs:2009:SCF:1577069.1577091,Koren091the}. In its basic form, matrix factorization profiles both items and users by vectors of factors inferred from rating patterns. High correspondence between item and user factors leads to a recommendation. Although matrix factorization is very popular because of its proven qualities, some shortcomings remain. One of these is the fact that the model generated by MF is static. Once it has been generated, the model delivers recommendations based on a snapshot of the incoming ratings frozen at the beginning of the generation. To take into account the missing ratings (those arrived after the last model generation), the model has to be computed periodically. However, it is not realistic to carry it out frequently, because of the high cost of model recomputation. Therefore, the quality of the recommendations will decrease gradually until a new model is computed.

In real-world context where new ratings happen continuously, users profile evolve dynamically. Consider, for instance, a costumer of an online music-store looking for good pop songs. He asks the application for some recommendations and the system proposes to him a short list of songs with high probability of interest (based on the latest available model). The costumer selects and rates the songs he already knows or he just listened to, and asks for new recommendations. Since the preferences of the customers evolve accordingly to the songs they have listened to, it is important to be able to integrate the new ratings for the subsequent recommendations. Otherwise, the accuracy of these recommendations will be low.
Online shops attempt to keep their customers loyalty and thus search to better satisfy them by relevant recommendations. This accounts for all attention brought to the evolution of user preferences. Indeed it has been claimed that even an improvement as small as 1\% of the accuracy leads to a significant difference in the ranking of the "Top-K" most recommended items for a user ~\cite{netflix-forum::how_useful_is_a_lower_RMSE,Dror:2011fk}.

In this paper, we propose a solution that reduces the loss of quality of the recommendations over time. It combines clustering, matrix factorization and bias adjustment~\cite{citeulike:4563135,Takacs:2008:IVM:1722149.1722155}, in order to startup with a high quality model. The biases are continuously updated with the new ratings, to maintain a satisfactory quality of recommendations for a longer time. Our solution is based on the observation that the rating tendency of a user is not uniform, and can change from one set of items to another. A set of biases is then associated to each user, one bias for each set of similar items. Thus, the integration of a new rating is provided by recomputing a local user bias (a bias of a user for a specific cluster of items), which may be done with a very low computation cost.

Our approach improves the scalability of recommender systems by reducing the frequency of model recomputations. The experiments we conducted on the Netflix dataset and the largest MovieLens dataset confirmed that our technique is well adapted for dynamic environments where ratings happen continuously~\cite{netflix-prize,grouplens}. The cost of the integration of new ratings is very low, and the quality of our recommendations does not decrease very fast between two successive matrix factorizations.
Also our idea of refining the user biases is orthogonal to the factorization models. It can be used in fully-fledged models with weights, temporal dynamics and so on ~\cite{Koren:2010:CFT:1721654.1721677,Koren:2009:MFT:1608565.1608614,Takacs:2008:IVM:1722149.1722155,Bell:2007:MRM:1281192.1281206}.

The remainder of this paper is organized as follows. In Section \ref{RS} we present some preliminary notions and requirements. Section \ref{CbMF} details our cluster-based matrix factorization solution. In Section \ref{Evaluation}, we present an experimental analysis of our proposal. Section \ref{Related_work} summarizes the related work, and Section \ref{Conclusion} concludes the paper.
\section{Preliminaries}
\label{RS}
This section defines the prediction problem and describes the matrix factorization technique on which our work is based. It also outlines the main requirements considered for the design of our system.

\subsection{Prediction issue}
The purpose of recommender systems is to predict the interest of a user for a given item, i.e. to determine how much the user would like the item. Most of the time, this interest is represented by numerical values from a fixed range. A set of interfaces, e.g. widgets, are often used to allow the users to rate the items. The ones used to enter ratings at a 1-to-5 star scale are still very popular on the web.

The prediction problem can be defined as follows.
Consider a set $U$ of users and a set $I$ of items. User ratings can be seen as tuples $(u, i, r_{ui}, t_{ui})$, where $u$  denotes a user, $i$ denotes an item, $r_{ui}$ the rating of user $u$ for the item $i$, and $t_{ui}$ is a timestamp. We assume that a user rates an item at most once.

The problem is to predict the future ratings such that the difference between an estimated rating $\hat{r}_{ui}$ and its true value $r_{ui}$ is the lowest possible. In order to build the estimator, the set of existing ratings is split in two parts: the first part is used for the training step and the second part for the evaluation of the accuracy of the estimator.

The quality of a recommender system can be decided on the accuracy of its predictions. The Root Mean Square Error (RMSE), which computes the root of the mean of the squarred difference between the predictions and true ratings, is one of the most widely used metric for the evaluation of recommender systems since the Netflix Prize \cite{Herlocker:2004,Su:2009:SCF:1592474.1722966,netflix-prize}. In this paper we use the RMSE metric to compare our proposition to traditional (static) systems.
\begin{equation}
    \label{RMSE-equation}
    RMSE = \sqrt{\frac{1}{n} \sum_{u,i}\left(r_{ui}-\hat{r}_{ui}\right)^2}
\end{equation}
\noindent
where $n$ is overall number of ratings. The lower the RMSE, the better is the prediction.

\subsection{Matrix factorization}
\label{RS_MF}
In the recommender systems using matrix factorization, the ratings are arranged into a sparse matrix $R$. The columns of $R$ represent the users where its rows represent the items. The value of each not empty cell $c_{ui}$ of $R$, corresponding to user $u$ and item $i$, is a pair of values $(r_{ui}, t_{ui})$. $r_{ui}$ is the rating given by $u$ for the item $i$ at time $t_{ui}$. An empty, i.e. missing, cell $c_{ui}$ in $R$ indicates that user $u$ has not yet rated item $i$. Hence, the task of recommender system is to predict these missing rating values. The table below represents such a matrix.

\begin{center}
  \begin{tabular}{c||c|c|c|c|c}
	  & $u_{1}$ & $u_{2}$ & \ \ \ldots\ \ \  & $u_{n}$ \\ \hline \hline
	  $i_{1}$ & 3 &  & \ldots & 1 \\ \hline
	  $i_{2}$ &  & 2 & \ldots & 5 \\ \hline
	  $i_{3}$ & 1 &  & \ldots &  \\ \hline
	  $i_{4}$ &  &  & \ldots &  \\ \hline
	  \vdots & \vdots & \vdots & \vdots & \vdots \\ \hline
	  $i_{m}$ &  & 4 & \ldots & 2 \\ \hline
	\end{tabular}
\end{center}

In its basic form (Basic MF), matrix factorization techniques try to capture the factors that produce the different rating values. They approximate the matrix $R$ of existing ratings as a product of two matrices:
\begin{equation}
    \label{MF-approximation-equation}
    R = P \cdot Q
\end{equation}

$P$ and $Q$ are called matrices of factors since they contain vectors of factors for the profiling of the users and the items, respectively. These matrices of factors are much more smaller than $R$. Thus, we gain in dimension while getting predictive ratings simply by the following formula
\begin{equation}
    \label{mf-prediction-equation}
    \hat{r}_{ui} = p_{u} \cdot q_{i}^{T}
\end{equation}
where $p_{u}$ and $q_{i}$ are the vectors of  factors, respectively in $P$ and $Q$, corresponding to user $u$ and item $i$.

In practice, it is very difficult to obtain exactly $R$ with the product of $P$ and $Q$. Usually, some residuals remain. These latter constitute the error of prediction, i.e. its inaccuracy, which can be represented by a matrix $E$ of errors having the same size than $R$. So, the previous equation can be changed to
\begin{equation}
    \label{MF-equation}
    R = P \cdot Q + E
\end{equation}

We can see that the more the matrix $E$ is close to a zero matrix, the more accurate will be the prediction. The process of training looks for the better values of $P$ and $Q$ such that the matrix $E$ is the closest possible to a zero matrix. Thus, it tries to adjust all the values $e_{ui}$ of the matrix $E$ to zero using a stochastic gradient descent (SGD) algorithm. The SGD algorithm computes a local minimum where the total sum of error values is one of the lowest according to initial ratings. In other words, it tries to minimize as good as possible the sum of quadratic errors $\sum\limits_{ui} e_{ui}^2$ between the predictive ratings $\hat{r}_{ui}$ and the real ones $r_{ui}$. Errors are squared in order to avoid the effects of negative values in the sum, and increase the weights of abnormal values. The fact of minimizing $\sum\limits_{ui} e_{ui}^2$ amounts to minimize each $e_{ui}^2$.

We have $e_{ui} \stackrel{def}{=} r_{ui} - \hat{r}_{ui}$. By using the vectors of factors $p_{u}$ and $q_{i}$, we obtain that $e_{ui} \stackrel{def}{=} r_{ui} - p_{u} \cdot q_{i}^{T}$. If we denote by $K$ the number of considered factors, we can avoid overfitting the observed data by regularizing the  squared error of know ratings. Thus we have the next regularized sum of squared errors
\begin{equation}
    \label{regularized_squared_error}
    \sum_{ui} e_{ui}^2 = \sum_{ui} (r_{ui} - p_{u} \cdot q_{i}^{T})^2 + \beta \cdot (\left\|p_{u}\right\|^2 + \left\|q_{i}\right\|^2)
\end{equation}
$\beta$ is a regularization factor which serves to prevent large values of $p_{uk}$ and $q_{ki}$. More precisely, we have
\begin{equation}
    \sum_{ui} e_{ui}^2 = \sum_{ui} (r_{ui} - \sum_{k}^K p_{uk} \cdot q_{ki})^2 + \beta \cdot (\left\|p_{u}\right\|^2 + \left\|q_{i}\right\|^2)
\end{equation}
\noindent
Then to minimize the quadratic errors, in order to get better predictions, we compute the differential (i.e., the gradients) of the squared error $e_{ui}^2$ to determine the part of change due to each factor ($p_{uk}$ and $q_{ki}$):
\begin{equation}
    \label{differentiate_equation}
    \frac{\partial e_{ui}^2}{\partial p_{uk}} = -2 \cdot e_{ui} \cdot q_{ki}~,~~~~~~~~\frac{\partial e_{ui}^2}{\partial q_{ki}} = -2 \cdot e_{ui} \cdot p_{uk}
\end{equation}
We update $p_{uk}$ and $q_{ki}$ in the opposite direction of the gradients in order to decrease the errors and thus obtain a better approximation of the real ratings.
\begin{equation}
	\label{updating_factors-equation-1}
	p_{uk} \leftarrow p_{uk} + \lambda \cdot (2 \cdot e_{ui} \cdot q_{ki} - \beta \cdot p_{uk})
\end{equation}
\begin{equation}
	\label{updating_factors-equation-2}
	q_{ki} \leftarrow q_{ki} + \lambda \cdot (2 \cdot e_{ui} \cdot p_{uk} - \beta \cdot q_{ki})
\end{equation}
$\lambda$ is a learning rate. The SGD algorithm iterates on the equations \ref{regularized_squared_error}, \ref{updating_factors-equation-1} and \ref{updating_factors-equation-2} until the regularized sum of the quadratic errors in the equation \ref{regularized_squared_error} does not decrease any more. This process corresponds to the training step.

After this training, the predictions $\hat{r}_{ui}$ are computed through the products $p_{u} \cdot q_{i}^{T}$ of both vectors of factors. A sorting step allows to find the most relevant items to recommend to each user, i.e. the items with the greatest product values.

\subsection{Biased MF}
\label{RS_BMF}
Several improvements to the above matrix factorization technique are proposed in the literature. One of these assumes that much of the observed variations in the rating values is due to some effects associated with either the users or the items, independently of any interactions ~\cite{Takacs:2009:SCF:1577069.1577091,Koren:2009:MFT:1608565.1608614,citeulike:4563135}. Indeed, there are always some users who tend to give higher (or lower) ratings than others, and some items may be higher (or lower) rated than others, because they are widely perceived as better (or worse) than the others. Basic MF can not capture these tendencies, thus some biases are introduced to highlight these rating variations. We call such techniques \textit{Biased MF}. The biases reflect users or items tendencies. A first-order approximation of the biases involved in rating $r_{ui}$ is as follows:

\begin{equation}
    \label{biases-equation}
    b_{ui} = \mu + b_{u} + b_{i}
\end{equation}

\noindent
$b_{ui}$ is the global effect of the considered biases, it takes into account users tendencies and items perceptions. $\mu$ denotes the overall average rating (for all the items, by all the users). $b_{u}$ and $b_{i}$ indicate the observed deviations of user $u$, respectively item $i$, from the average. Hence, the equation \ref{mf-prediction-equation} becomes
\begin{equation}
    \label{mf_biased-prediction-equation}
    \hat{r}_{ui} = p_{u} \cdot q_{i}^{T} + \mu + b_{u} + b_{i}
\end{equation}
Since biases tend to capture much of the observed variations and can bring significant improvements, we consider that their accurate modeling is crucial~\cite{citeulike:4563135,Koren091the}. As for the factors $p_{uk}$ and $q_{ki}$ (equations \ref{updating_factors-equation-1} and \ref{updating_factors-equation-2}), the biases have to be refined through a training step using the following equations:

\begin{equation}
	\label{updating_bias_bi}
	b_{i} \leftarrow b_{i} + \lambda \cdot (2 \cdot e_{ui} - \gamma \cdot b_{i})
\end{equation}
\begin{equation}
	\label{updating_bias_bu}
	b_{u} \leftarrow b_{u} + \lambda \cdot (2 \cdot e_{ui} - \gamma \cdot b_{u})
\end{equation}
where $\gamma$ is a regularization factor. It plays the same role than $\beta$ in equations \ref{updating_factors-equation-1} and \ref{updating_factors-equation-2}.

\subsection{Dynamicity and performance requirements}
\label{performance_requirements}
\paragraph{Dynamicity problem}
\noindent
As told above, once a model is carried out, it remains static unless a new MF is computed. In real-world context, where new ratings happen continuously, the user interests evolve dynamically. Thus, the accuracy of the predictions decreases gradually and the computed profiles become obsolete after some time, since they do not take into account the new additions of ratings. To face this problem, recommender systems must regularly recompute their models, which represents an expensive task in terms of computation time. Hence, the dynamicity problem can be defined as follows: how to integrate the new ratings in the predictions without recomputing the model? The goal is to maintain the accuracy of the predictions at a good level and postpone as far as possible the recomputation of the model.

\vspace{.3cm}
We present in the following some important
requirements that the solutions for the dynamicity problem must satisfy.

\paragraph{Recommendation quality}
\noindent
Assuming some fixed sets of users and items. We consider users continuously asking for items, and rating them. For instance, a user asks for a short list of items with high probability of interest (i.e. high predicted rating), then selects and rates some of them, and so on. In such online recommendation scenario, the user expects the recommended items to be of high interest. We measure the quality of service in terms of the Root Mean Square Error (RMSE) between the predicted and the real ratings. We express the user requirement for quality, as a constraint on the RMSE, which value must be greater than a given threshold $\epsilon$.
\begin{equation}
	RMSE < \epsilon
\end{equation}

\paragraph{Response time}
\noindent
Another requirement for online recommendation is the response time tolerated by the end users. When a user asks for a recommendation, he expects to receive it almost immediately. Such requirement for online user demand is usually described by an upper bound along with a ratio of appliance \cite{tpcc}: 90\% of the demands must be served in less than 5 seconds. This response time constraint forces us (1) to generate the model in advance, in order to anticipate the future demands, and (2) to limit the computational cost needed for the integration of the ratings arrived after the model generation.
%

\vspace{.3cm}
We can summarize the performance requirements into the following challenge: design a recommendation system which provides sufficient quality, when generating the "top-quality" model takes a long time, when the predictions quality is decreasing over time, featuring fast recommendation delivery on user demand, and reducing the overall computation cost.

Our solution to tackle this challenge is based on the following process:
\begin{enumerate}
\item Combine clustering, MF and bias adjustment, to take into account the specificity of each user and start with a high quality model.
\item Continuously update the biases (with a low computation cost), in order to maintain as long as possible the quality of the predictions at satisfactory level.
\item Estimate the forthcoming quality loss, and regenerate a high quality model when the quality loss becomes important.
\end{enumerate}

In the following we will detail our solution for the first two points. The estimation of the forthcoming loss of quality of recommendations is beyond the scope of this paper. Some recent work~\cite{Jambor:2012:UCT:2187836.2187839}, shows interesting directions that we plan to study in the future work.

\section{Dynamic recommendations}
\label{CbMF}
As told above, we focus on dynamic contexts where new ratings are continuously produced. In such case, it is not possible to have an up to date model, due to the incompressible time needed to compute the recommendation model. At least, the ratings produced during the model computation will be missing.
After each generation of a new model, the situation can degrade quickly enough since the number of non processed ratings may increase very fast. Then, a growing loss of quality can be observed in the recommendations, as long as the static model is used.

To tackle this problem, our model relies on biases which are among the most overlooked components of recommender models~\cite{Koenigstein:2011:YMR:2043932.2043964}.
Biases allow to capture a significant part of the observed rating behavior.
We combine global user biases with local user biases. The local user biases allow to refine user's tendency on small sets of items, whether the global biases capture the general behaviors of the users. To be accurate (i.e., allow good predictions of users ratings), local biases need to be computed on sets of similar items. These sets can be obtained by a clustering step, as proposed in our approach. The global user biases guarantee a certain stability. In case where the local user bias has not enough information (ratings), the global user bias plays a role of balance. It ensures, in the worst case, that user's tendency will follow her general behaviour.

In the following, we first highlight the importance of our clustering, then we detail our proposed solution which combines global biases and cluster-based local biases. And lastly, we present the algorithm that integrates the new ratings by adjusting the local biases in the recommendation model.

\subsection{Why clustering ?} 
We argued above that the accuracy of local user biases depends on the degree of similarity between the items in each set (i.e. cluster). This section formalizes the relation between the similarity of a set of items and the variance of users biases. We show that the more similar are the items in each cluster, the more the variance of the local user biases is small. A smaller variance means a lower prediction error, thus a more accurate recommendation.

Let $U$ be a set of users, $I$ a set of items, $r_{ui}$ a rating of a user $u \in U$ for an item $i \in I$, and $\mu$ the overall average of rating. Consider $I_u \subset I$, the set of items rated by a user $u$, then the bias $b_u$ of the user $u$ is defined as follows:
\begin{equation}
    \label{user_bias_definition_1}
    b_u = \frac{1}{card(I_u)} \sum_{i \in I_u} \left(r_{ui}-\mu\right)
\end{equation}
For a given item $i \in I_u$, the local deviation of the user $u$ relative to the overall average of rating $\mu$ is:
\begin{equation}
    \label{user_bias_local_deviation}
    b_{ui} = r_{ui}-\mu
\end{equation}
Then, equation \ref{user_bias_definition_1} can be simplified as:
\begin{equation}
    \label{user_bias_definition_2}
    b_u = \frac{1}{card(I_u)} \sum_{i \in I_u} b_{ui}
\end{equation}

To measure the user bias variation, we compute for each user $u$ her bias variance $Var_u$ as follows:
\begin{equation}
    \label{user_bias_variance_1}
    Var_u = \frac{1}{card(I_u)} \sum_{i \in I_u} {\left(b_{ui}-b_u\right)}^2
\end{equation}

\noindent
Then, equations \ref{user_bias_local_deviation}, \ref{user_bias_definition_2} and \ref{user_bias_variance_1} lead to the following formula:
\begin{equation}
    \label{user_bias_variance_2}
    Var_u = \frac{1}{{card(I_u)}^3} \sum_{i \in I_u} {\left({\sum_{j \in I_u} {(r_{ui}-r_{uj})}}\right)}^2
\end{equation}

To compute the variance, the user must have at least two ratings. Then, the variance can be bound as shown in the following equation:
\begin{equation}
\begin{split}
    \label{user_bias_variance_3}
    Var_u \leq \frac{1}{{2}^3} \sum_{i \in I_u} {\left({\sum_{j \in I_u} {(r_{ui}-r_{uj})}}\right)}^2 \\
    \leq \frac{1}{8} {\left({\sum_{i \in I_u} \sum_{j \in I_u} {\left|r_{ui}-r_{uj}\right|}}\right)}^2
\end{split}
\end{equation}
Then, considering all the users we obtain:
\begin{equation}
    \label{total_user_bias_variances}
    0 \leq \sum_{u \in U} Var_u \leq \frac{1}{8} {\left(\sum_{u \in U} \sum_{i \in I_u} \sum_{j \in I_u} {\left|r_{ui}-r_{uj}\right|}\right)}^2
\end{equation}

\noindent
\emph{Measuring the dissimilarity of items.}\\
 Consider two items $(i,j) \in I^2$, and let $U_{ij} \subset U$ be the set of users having rated both them. The dissimilarity of the items $i$ and $j$ can be measured according to the difference of the ratings $r_{ui}$ and $r_{uj}$ given to them by each user $u$.
Hence, we define the dissimilarity of two items $(i,j) \in I^2$ as follows:
\begin{equation}
    \label{squared_item_dissimilarity}
    dissim_{ij} =  \sum_{u \in U_{ij}} {\left|r_{ui}-r_{uj}\right|}
\end{equation}
$dissim_{ij}$ tends to zero when all the users in $U_{ij}$ have close ratings for both items. The sum of the dissimilarities of all the couples of items is:
\begin{equation}
    \label{total_squared_item_dissimilarities}
    \sum_{(i,j) \in I^2} dissim_{ij} = \frac{1}{2} \sum_{i \in I} \sum_{j \in I} \sum_{u \in U_{ij}} {\left|r_{ui}-r_{uj}\right|}
\end{equation}
Since $dissim_{ij} = dissim_{ji}$, we divide by 2 the sum in the right part of the previous equation.

\noindent
Equations \ref{total_user_bias_variances} and \ref{total_squared_item_dissimilarities} lead to the following ascertainment on the dissimilarity of the items and the user bias variances: 
\begin{equation}
    \label{dependency}
    0 \leq \sum_{u \in U} Var_u \leq {\left(\sum_{(i,j) \in I^2} dissim_{ij}\right)}^2
\end{equation}
For a given set $I$, the less dissimilar (i.e., more similar) are the items (i.e. $\sum_{(i,j) \in I^2} dissim_{ij} \rightarrow 0$), the less varying are the user biases (i.e. $\sum_{u \in U} Var_u \rightarrow 0$). In other words, the users tend to have uniform behaviours on such a set of similar items. So, defining a bias for each user, and on each set of similar items, leads to a small variance in the local biases and, consequently, a good accuracy in the predictions. The clustering step is then a crucial part of our approach.

\subsection{The CBMF model}
Our cluster-based matrix factorization model (CBMF) is based on the observation that many users usually tend to underestimate (or overestimate) the items they rate. A user may have a tendency to rate above (or beyond) the average. We aim to quantify such tendency. A simple way to take it into account is to assign a single bias per user (as shown in section \ref{RS_BMF}). However, we observed that users tendency is generally not uniform: it can change from one item to another. For some sets of items, a user can tend to rate close to the average. While for some other items (e.g., those she really likes/dislikes), the user fails to rate objectively, either using extreme ratings, or keeping moderated ratings.

To take into account this discrepancy, we define several biases per user, instead of a
single one. We assign one bias $b^C_u$ for each user $u$ and each set $C$ of similar items. We rely on existing clustering techniques to group similar items
together. We expect that handling finer-grained biases will lead to more accurate recommendation. In our context, we consider that the only
known information about the items is their ratings. Additional information or properties of the items could be considered in the clustering phase, but this is not the purpose of our work and remains out of the scope of this paper.
Once the clusters are built, we assign a vector of biases to each user. One bias for each group of items. Then, we apply our matrix factorization (CBMF) on the ratings to generate the recommendation model.

Thus, we come down to observe local ratings variation in place of a single global ratings variation as used in previous approaches~\cite{citeulike:4563135,Koren:2009:MFT:1608565.1608614,Takacs:2008:IVM:1722149.1722155}.
We derive the bias $b^C_u$ of a user $u$ in a cluster $C$ from the ratings of the items contained in this cluster. For each rated item $j \in C$ (by user $u$), we define the deviation $b_u^j$ of user $u$ for this item as the difference between her rating for $j$ and the average rating $\mu^{C}$ of all the users for the items in cluster $C$: $b_u^j = r_{uj} - \mu^{C}$.
The local bias of the user $b_u^C$, at the level of the cluster, is obtained by taking her average deviation as shown in equation \ref{bias_per_group-equation}.
\begin{equation}
	\label{bias_per_group-equation}
	b^C_u = \frac{1}{|C|} \sum_{j \in C} r_{uj} - \mu^{C} ~~~~\forall{j \in C, ~s.t.~r_{uj} > 0}
\end{equation}
Let us remind that we have used zero as a cell value in matrix $R$ to represent the missing ratings. Therefore the ratings are superior to zero (between 1 and 5, generally). This explains the condition $r_{uj} > 0$ in equation \ref{bias_per_group-equation}.

In our approach we try to find the best trade off between local and global biases. The gap between these biases is moderated by the relative number of ratings the user have in each cluster. We define $\delta^C_u$ as the weighted difference between the local bias $b^C_u$ of user $u$ in the group of items $C$ and his global bias $b_u$. In equation \ref{weighted-difference}, $n^C_u$ denotes the number of ratings user $u$ has in the group of items $C$, and $n_u$ denotes his/her total number of ratings.
\begin{equation}
    \label{weighted-difference}
    \delta^C_u = \frac{n^C_u}{n_u} \cdot (b^C_u - b_u)
\end{equation}

Thus, our prediction formula is the following:
\begin{equation}
    \label{cbmf-prediction-equation}
    \hat{r}_{ui} = p_{u} \cdot q_{i}^{T} + \mu^{c(i)} + \delta^{c(i)}_u + b_u + b_{i}
\end{equation}
\noindent
where $c(i)$ denotes the group/cluster to which the item $i$ belongs
and $b_{i}$ represents the observed deviation of item $i$.
From this, the regularized global sum of squared errors becomes:

\begin{equation}
\begin{split}
    \label{our_model_regularized_squared_error}
    \sum_{ui} e_{ui}^2 = \sum_{ui} (r_{ui} - (p_{u} \cdot q_{i}^{T}) + \mu^{c(i)} + \delta^{c(i)}_u + b_u + b_{i})^2\\
    + \beta \cdot (\left\|p_{u}\right\|^2 + \left\|q_{i}\right\|^2 + {\delta^{c(i)}_u}^2 + {b_u}^2 + {b_{i}}^2)~
\end{split}
\end{equation}
As the global biases $b_u$ and $b_i$, the local biases $b^{c(i)}_u$ have to be refined through their weighted differences $\delta^{c(i)}_u$ using the formula:
\begin{equation}
	\label{updating_local_bias}
	\delta^{c(i)}_u \leftarrow \delta^{c(i)}_u + \lambda \cdot (2 \cdot e_{ui} - \gamma \cdot \delta^{c(i)}_u)
\end{equation}

The algorithm \ref{algo_cluster-based_mf} details the steps of our CBMF process.
In line 1, the clustering of the input ratings is processed. Line 2 computes the initial bias value of each item, the global bias of each user and his local biases. The initial set of weighted differences $\left\{\delta^{C}_u\right\}$ between local and global biases is also computed at this step. From them we can deduce the user local biases. Line 3 initializes the matrices of factors $P$ and $Q$. This is done with random low values.
Lines 4 to 11 correspond to the main part of the learning process. At each iteration (lines 5 to 10), the error of prediction $e_{ui}$ is computed for each rating. Then, the matrices of factors, the biases (global and local ones) are adjusted accordingly (lines 7 to 11), using equations \ref{updating_factors-equation-1}, \ref{updating_factors-equation-2}, \ref{updating_bias_bi}, \ref{updating_bias_bu}, and \ref{updating_local_bias}.
Line 13 measures the global error as indicated in equation \ref{our_model_regularized_squared_error}. The training process ends when the regularized global squared error does not decrease any more or when the maximum number of iterations is reached.

\begin{algorithm}
	\label{algo_cluster-based_mf}
	\SetAlgoLined
	\KwData{$N_c$: number of clusters, $\mathbf{R}$: matrix ${\mathbb N}^{m*n}$ of ratings, $K$: number of factors to consider, $\lambda$, $\beta$ and $\gamma$}
	\KwResult{$P$, $Q$, $\mu = \left\{\mu^C\right\}$, $b_{i}$, $b_u$ and $\left\{\delta^{C}_u\right\}$, $~C \in C_1, C_2, ... C_{N_c}$}
	\BlankLine
	Compute the clusters $C_1, C_2, ... C_{N_c}$ from the input data $\mathbf{R}$\;
	For each item $i$ and each user $u$, calculate the biases $b_{i}$, $b_{u}$ and $\left\{\delta^{C}_u\right\}$, $~C \in C_1, C_2, ... C_{N_c}$\;
	Initialize the matrices $P$ and $Q$, respectively of dimensions $m*k$ and $k*n$\;
	\Repeat{terminal condition is met}{
		\ForEach{ $r_{ui} \in \mathbf{R}$ }{
	      Compute $e_{ui}$\;
				\For{$k\leftarrow 1$ \KwTo $K$}{
						Update $p_{uk} \in P$, $q_{ki} \in Q$\; 
				}
				Update $b_{i}$ and $b_{u}$\;
				Update also $\delta^{c(i)}_u$\; 
		}
	  Calculate the global error $\sum_{r_{ui}>0} e_{ui}^2$\; 
	}
	\Return{$P$, $Q$, $\mu = \left\{\mu^C\right\}$, $b_{i}$, $b_{u}$, $\left\{\delta^{C}_u\right\}$, $~C \in C_1, C_2, ... C_{N_c}$}
	
		\caption{Cluster-based MF algorithm}
\end{algorithm}

\subsection{Integration of incoming ratings}
After the generation of the recommendation model, the incoming ratings continue to be added to the ratings matrix $R$. Their integration in the model is done simply by adjusting the local user biases.

Hence, the importance of local biases. Indeed, the top-K item recommendation is constituted generally of items from different clusters (in our experimentations, for three clusters, we observed that 58.47\% of the users of Netflix have at least two clusters represented in their top-5, and 55.12\% for MovieLens). When we adjust the local user biases with the new ratings, the recommendations can be affected in the composition of the recommended list of items or in the ranking (top-K) of these items.

Let us denote by $V$ the set of known ratings in $R$, including the newly added ones.
\begin{equation}
	V = \left\{r_{ui} \in R / ~u \in U, i \in I~~and~~r_{ui} > 0 \right\}
\end{equation}
\noindent
where $U$ and $I$ are the sets of referenced users and items, respectively.
Then, we denote by $V(u,.)$ the set of all known ratings of a given user $u \in U$.
\begin{equation}
	V(u,.) = \left\{r_{ui} \in V, \forall i \in I\right\}
\end{equation}

The subset of ratings of user $u$ in the cluster $c(i)$ to which a specific item $i$ belongs is denoted by $V(u, c(i))$.
\begin{equation}
	V(u, c(i)) = \left\{r_{uj} \in V(u,.) / j \in c(i)\right\}
\end{equation}

The bias adjustment done when a new rating $r_{ui}$ is obtained, requires only the ratings in $V(u, c(i))$.
A gradient descent is performed to update the local bias of user $u$ in the cluster $c(i)$, using equation \ref{updating_local_bias}.
The algorithm \ref{algo_online-updating} details the steps of the ratings integration process. As in Algorithm \ref{algo_cluster-based_mf}, the training process ends when the regularized global squared error does not decrease any more or when the maximum number of iterations is reached.

\begin{algorithm}
	\label{algo_online-updating}
	\SetAlgoLined
	\KwData{$P$, $Q$, $V(u, c(i))$, $b_{i}$, $b_{u}$, $\delta^{c(i)}_u$, $\lambda$, $\beta$ and $\gamma$}
	\BlankLine
	\Repeat{terminal condition is reached}{
			\ForEach{ $r_{uj} \in V(u, c(i))$ }{
					Compute $e_{uj}$\;
					Update $\delta^{c(i)}_u$\;
			}
			Calculate the global error $\sum_{{r_{uj}>0}}{e_{uj}^2}$\;
	}	
	\caption{Incoming ratings integration algorithm}
\end{algorithm}

\subsection{Complexity analysis} 
\label{Formal_evaluation}

The cost of our cluster-based matrix factorization solution (Algorithm \ref{algo_cluster-based_mf}) can be separated in two parts: the cost of matrix factorization and the cost of the clustering step. The time complexity of the training of the whole model (matrix factorization) is $O(\left|V\right| \cdot k \cdot t)$, where $V$ denotes the set of known ratings, $k$ is the number of factors and $t$ the maximum number of iterations. The time complexity of the clustering step depends on the chosen clustering algorithm.

When additional information on the items is available (metadata on the items), it may be used for clustering~\cite{Koenigstein:2011:YMR:2043932.2043964,Ziegler:2008:ECT:1460172.1460177}. Such methods can greatly reduce the clustering execution time. If no metadata is available, there are still many possible clustering techniques, only based on item ratings, each one having its own cost: projected K-means, PDDP and so on~\cite{Kogan:2006:GMD:1121696,Kogan:2007:ICL:1214086,Sun:2010:KPC:1844770.1845516}.

The strength of our technique lies in the low computation cost needed for the integration of the ratings received after the generation of the model. So that the integration can be done on the fly and the loss of quality of the recommendations slowed. The time complexity of the integration of a new rating $r_{ui}$ is $O(\left|V(u, c(i))\right| \cdot t)$. Note that in the worst case this cost is equal to $O(\left|V(u,.)\right| \cdot t)$, when all the ratings of the considered user are related to the same group of items. Let us stress that $V(u,.)$ is usually small. For instance, for Netflix the average size of $V(u,.)$ is 200~\cite{netflix-prize}. The more the user ratings are distributed in different groups, the more the cost of updating the user bias is small. Still for Netflix, we have 98.4, 48.7, and 70.3 ratings in average per user with our three clusters of items.

\section{Experimental evaluation}
\label{Evaluation}

In Section \ref{performance_requirements} we proposed to enhance the widely used MF model, coupling it with two techniques that tend to improve the quality of predictions: the preliminary clustering of the ratings before factorization, and the final adjustment of the predicted ratings using biases.
This section presents the experiments we settled, in order to validate our approach.
We remind that our approach consists of generating a high quality recommendation model based on incoming ratings. Then, we use that model for recommending items, as long as possible (provided that quality remains sufficient) up to next generated model is ready, and so on. Thus, the quality of our approach depends on two factors (i) the initial quality of the generated model, and (ii) the loss of quality over time. Accordingly, we validate each factor independently, proceeding in two separated steps. Step 1 focuses on the initial quality of the model that has just been generated. Step 2 focuses on the loss of quality, of our approach, over time.

\setcounter{paragraph}{0}
\paragraph{\textbf{Step 1: Validation of the initial quality}}
We plan to show that our model yields good initial predictions compared to other commonly used models. We setup a fully informed environment, meaning that the model is aware of all the ratings that precede the prediction. This environment is optimal since it provides the maximal input to the model generation. Although this environment is rarely met in practice (it implies that no new ratings have occurred during the model generation), it ensures the most favorable conditions for every model. Thus it allows us comparing several models when they expose their best strength.
Our objective is to quantify the quality of our model that combines factorization with clustering and bias adjustment. To this end, we compare the accuracy of our model with two commonly used models: (i) the MF alone, and (ii) the biased MF (see Section \ref{RS_BMF}). Note that, we do not compare our solution with the case of MF preceded by clustering without bias adjustment, since clustering does not improve the accuracy directly in its own. Actually, clustering allows finer biases (one bias per cluster), which in turns yields better accuracy.

\paragraph{\textbf{Step 2: Validation of the loss of quality over time}}
In the second validation step, we check that the accuracy of prediction decreases over time after each factorization. This aims to justify the relevance of our investigation to provide predictions which accuracy lasts longer. Then, we will measure the benefits of our approach (continuous bias update, based on new ratings) for keeping up the accuracy of prediction longer than others. In other words, our solution should expose a smaller quality decrease (i.e. a flatter slope) than other solutions. In consequence, it will imply less frequent model re-regeneration, saving a lot of computation work.

\subsection{Implementation and Experimental setup}
We implemented our proposition in C++ and ran our experiments on a 64-bits linux computer (Intel/Xeon x 8 threads, 2.66 Ghz, 16 GB RAM). We used a LIL matrix structure to store the dataset of ratings. To cluster the items, we ran a basic factorisation with some iterations and a K-Means algorithm on the items factors.\newline
We made preliminary tests to calibrate the parameters of the model and the number of clusters: $\lambda = 0.001$, $\beta = 0.02$, $\gamma = 0.05$, $N_c = 3$. The $\lambda$, $\beta$, and $\gamma$ values are close to the ones suggested in~\cite{citeulike:4563135}. We limit training to 120 iterations at most and use 40 factors for both matrices $P$ and $Q$.
 

\subsection{Datasets}
\label{dataset}
We conduct the experiments on the Netflix dataset and the largest MovieLens datasets~\cite{netflix-prize,grouplens}. These datasets are very often used by the recommendation system community~\cite{Su:2009:SCF:1592474.1722966}. Table \ref{tab:1} shows their caracteristics.
\begin{table}
  \begin{center}
  	\caption{\label{tab:1} Caracteristics of the datasets}
		\begin{tabular}{|l|c|c|c|}
		 \hline
		  Size & \# of ratings & \# of users & \# of movies \\ \hline
		  MovieLens & 10M & 71,567 & 10,681 \\ \hline
		  Netflix & 100M & 480,189 & 17,770 \\ \hline
		\end{tabular}
	\end{center}
\end{table}
The ratings are represented by integers ranging from 1 to 5 for the Netflix dataset and real numbers for the one of Movielens.
Each dataset is ordered by ascending date.

\subsection{Initial quality}
The objective of this experiment is to compare the initial qualities
of the three models. We split the datasets into two parts : a training set
representing 98\% of the set of ratings and a test set which
keeps the rest (the 2\% most recent ratings to predict). So the test set
contains 1.88M ratings. This number of ratings is greater than the one
of the Netflix Prize which has 1.4M ratings~\cite{netflix-prize}.
Table \ref{tab:initial_quality} reports the different RMSE
errors obtained for the three models named Basic MF, Biased MF, and CBMF.
We remark that CBMF outperforms other models. It
reaches 1.12\% of improvement over the biased MF with the Netflix
dataset. Let us remind there, even an improvement as small as 1\% of the accuracy
leads to a significant difference in the ranking of the "Top-K" most recommended
items for a user ~\cite{netflix-forum::how_useful_is_a_lower_RMSE,Dror:2011fk}.
\label{exp:initial_quality}
\begin{table}
  \begin{center}
		\caption{\label{tab:initial_quality} Initial quality of the three models}
		\begin{tabular}{|l|c|c|c|}
		 \hline
		  Dataset & Basic MF & Biased MF & CBMF \\ \hline	
		  Movielens & 0.7743 & 0.7608 & \textbf{0.7578} \\ \hline 
		  Netflix & 0.9599 & 0.9312 & \textbf{0.9208} \\ \hline
		\end{tabular}
	\end{center}
\end{table}

\subsection{Large training sets improve quality}
\label{exp:size_impact}
The objectif of this experiment is to measure the quality of the model
according to the size of the training set.
We check the intuitive rule stating that the more ratings
we take as input, the best quality we get.

\begin{figure}
  \begin{center}
           \includegraphics[scale=0.5]{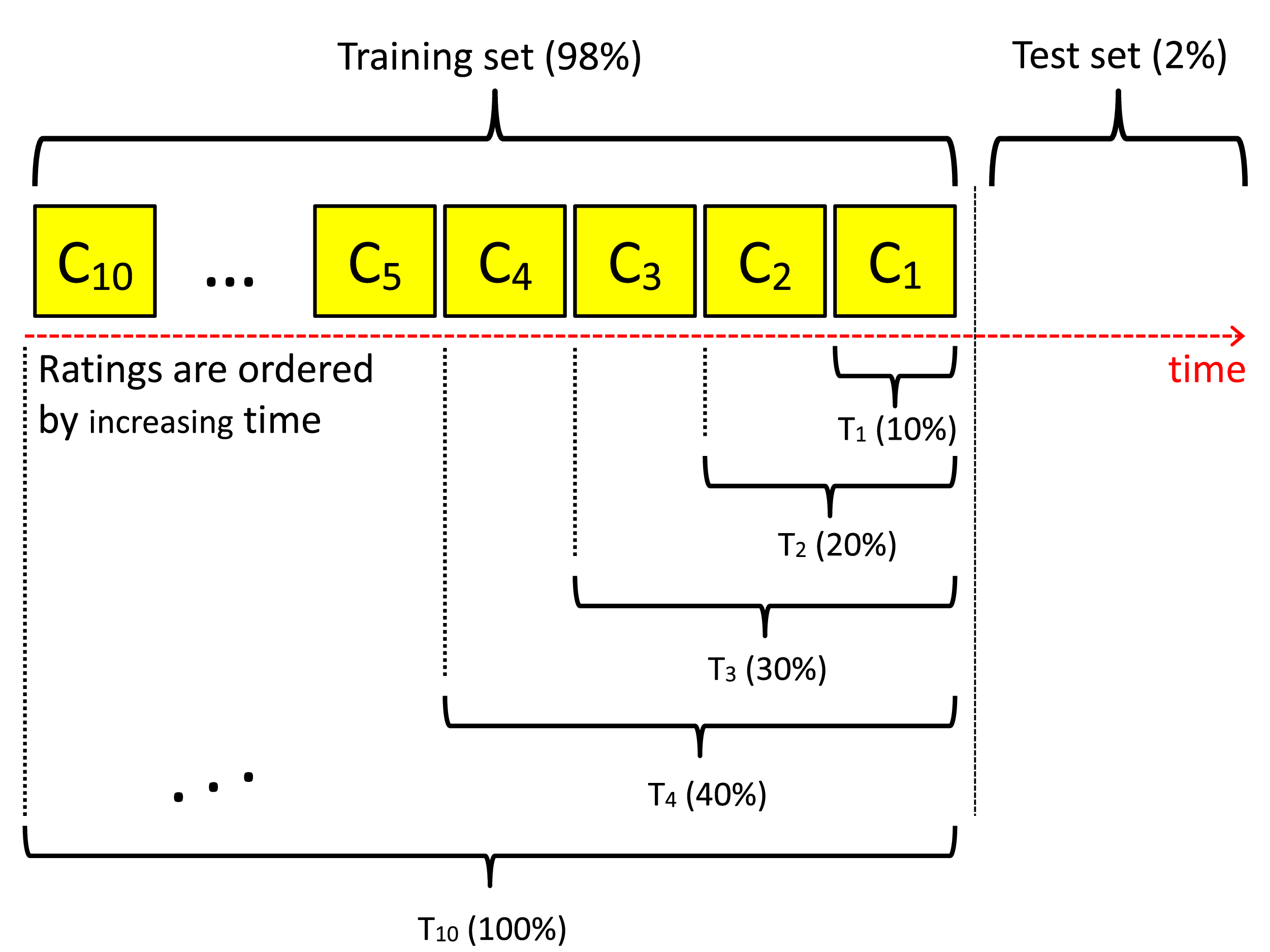}
          \caption{\label{fig:dec} Training sets partitioning}
  \end{center}
\end{figure}

To realize this experiment, we first sorted the ratings of each user according to their timestamps. Then, we split the training set (98\% of the initial dataset) into 10 chunks ($c_1$ to $c_{10}$) of equal size: 10\% each. Thus, the number of ratings of a user is almost the same in each chunk. From that, we generate 10 training sets ($T_1$ to $T_{10}$) of increasing size by assembling the chunks such that we always use the most recent ratings to generate the model.
More precisely, $T_1 = \left\{c_{10}\right\}$, $T_2 = \left\{c_9\right\} \bigcup \left\{c_{10}\right\}$, $T_3 = \bigcup_{i \in [8-10]} \left\{c_i\right\}$, ... $T_{10} = \bigcup_{i \in [1-10]} \left\{c_i\right\}$. (cf. Figure \ref{fig:dec}).

\begin{figure}
    \centering
    \subfigure[MovieLens]{\label{subfig_10M} \includegraphics[scale=0.5]{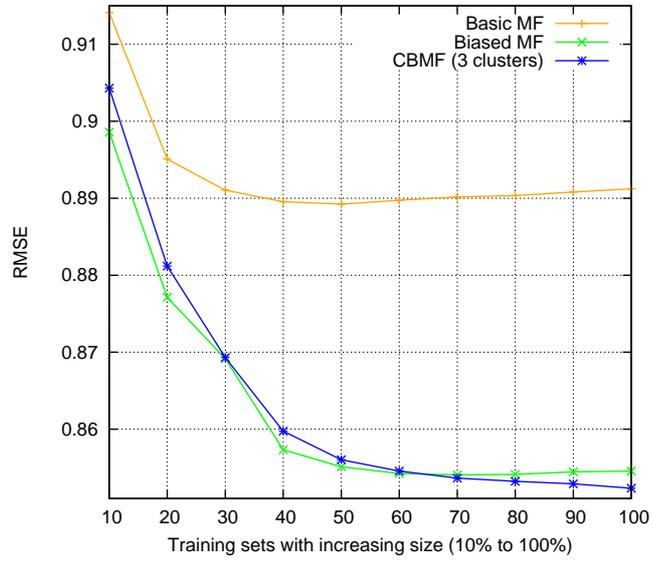}}
    \subfigure[Netflix]{\label{subfig_100M} \includegraphics[scale=0.5]{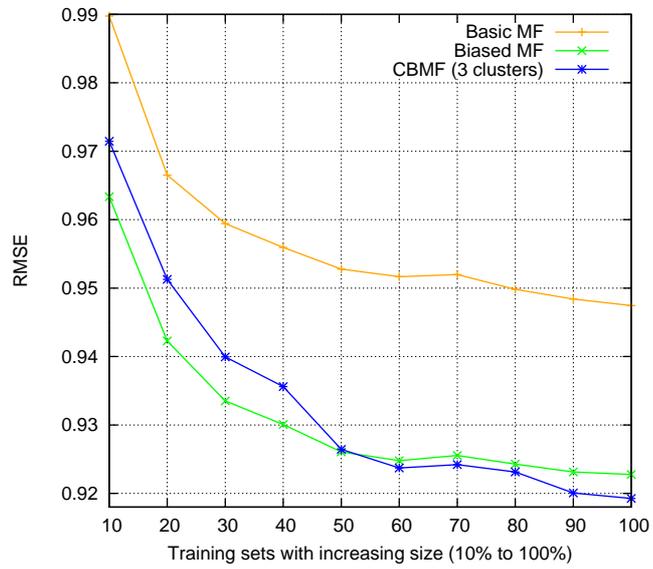}}
    \caption{\label{fig:size_impact} Quality improvement for increasing training sets sizes}
\end{figure}

Figure \ref{fig:size_impact} reports the RMSE evolution of the three models, for the two datasets: MovieLens (\ref{subfig_10M}), and Netflix (\ref{subfig_100M}).

We see that the three models are affected by the size of the training
set. The more ratings they have, the better quality they tend to
propose.  Table \ref{tab:progression} shows the quality improvements of these three models from $T_1$ to $T_{10}$.
The CBMF model shows 5.7\% and 6\% of quality improvements respectively for Netflix and MovieLens, thanks to
the finer-grained cluster-based bias adjustment.
This confirms the ability of local biases to better capture user tendencies over large training sets.
We observe on Figure~\ref{fig:size_impact} that on the range 10\%-60\%
(training sets $T_1$ to $T_6$), the Biased MF model outperforms the CBMF
model. Indeed, with the first training sets we do not much data to
compute enough discriminative clusters. Also the fact that the users do not have yet
rated a lot of items harms the local biases adjustment.

\begin{table}
  \begin{center}
	  \caption{\label{tab:progression} Percentage of quality improvement}
		\begin{tabular}{|l|c|c|c|}
		 \hline
		  Dataset & Basic MF & Biased MF & CBMF \\ \hline	
		  Movielens & 2.56 & 5.15 & \textbf{6.09} \\ \hline
		  Netflix & 4.46 & 4.39 & \textbf{5.67} \\ \hline
		\end{tabular}
	\end{center}
\end{table}

We also see different RMSE error ranges between the datasets. This difference between the RMSE errors is due to the data characteristics. For instance, the 10M MovieLens dataset has decimal ratings while the Netflix dataset uses only integer values. Adomavicius and Zhang mention this phenomenon in~\cite{Adomavicius:2012:IDC:2151163.2151166}. They point out consistent and significant effects of several data characteristics on recommendation accuracy.
Finaly, we note the importance of the biases. The basic MF suffers from that, it never catches up the other models whatever the dataset.

\subsection{Quantifying the need for online integration}
\label{exp:need_integration}
Basically, we need online integration when offline solutions fail to provide sufficient quality. The objective here is to measure the impact of missing ratings on the quality that offline models can deliver. We wonder to what extent the most up-to-date ratings influence the recommendation. Given a training set containing a fixed amount of ratings, we investigate the quality variation when the ratings become less and less recent.
Moreover, we target the \textquoteleft input intensive\textquoteright~ scenarios where a lot of new ratings are produced in a short period of time, thus million ratings are potentially missing. For instance Netflix company receives 4 million ratings per day~\cite{netflix-forum::beyond_the_5_stars_part1}. To reflect this, we must consider several millions of missing ratings in our experimentations. Therefore, we experiment only with the Netflix dataset which is the largest one, the MovieLens dataset does not have enough ratings to setup an enough number of missing ratings. Indeed 10M Movielens dataset does not match the experimental requirements, because we risk to reduce drastically the training set size, which becomes too small to produce meaningful results (i.e., few items are rated in both the test set and the training set).

We define the test set and the training set as follows. We keep in the
test set 10\% of the ratings, the most recent ones. The training set
contains the 90\% remaining ratings.
To better observe the impact of the delay on the RMSE, we balance the delay of each user.
More precisely, we order the ratings by arrival position, such that the $i^{th}$ ratings of any user
preceed the ${i+1}^{th}$ ratings of any of them, and so on.
%
%
We measure the evolution of the predition
quality along the ordered test set by computing the RMSEs over a sliding
window of 200K ratings as size. So that two consecutive windows
share the half of their ratings (for smother results).

\begin{figure}
  \begin{center}
           \includegraphics[scale=0.5]{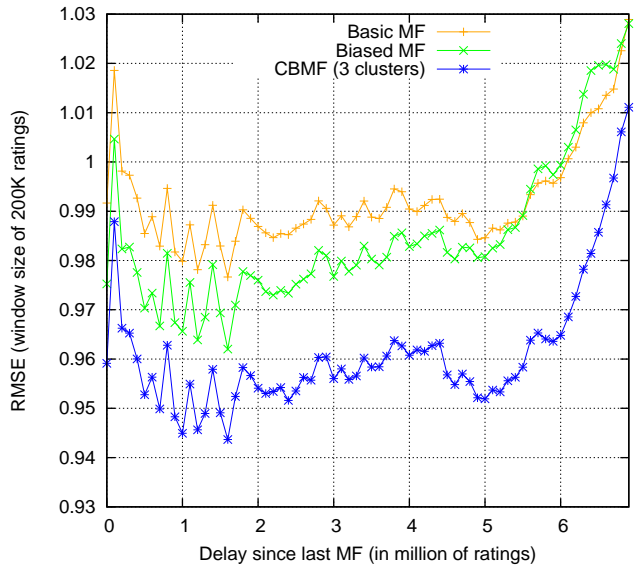}
          \caption{\label{fig:offline} Offline quality (RMSE value) with increasing delays (in million of ratings)}
  \end{center}
\end{figure}

Figure \ref{fig:offline} shows the evolutions of the prediction quality for the
three models : Basic MF, Biased MF, and CBMF.

Figure \ref{fig:offline} shows that the error is increasing with the
number of missing ratings. We observe a 5\% RMSE increase for long delays (from 5M to 7M missing ratings).
Such quality loss might not be acceptable for recommendation
systems. This confirms the need for online integration.
%


\subsection{Robustness to time of our online integration model}
\label{exp:robustness}
The goal is to show that our model is robust to time, i.e., it still
yields good quality predictions even when many ratings have been
produced since the last factorization.  Using the same training and
test sets, we now take into account the missing ratings to adjust on the fly
the local users' biases (cf. Algorithm \ref{algo_online-updating}).
More precisely, we sequentially scan the test set and consider the ratings one by one. For each rating,
we calculate the prediction error, then we immediatly integrate the rating in order to
improve the next predictions.

\begin{figure}
  \begin{center}
           \includegraphics[scale=0.5]{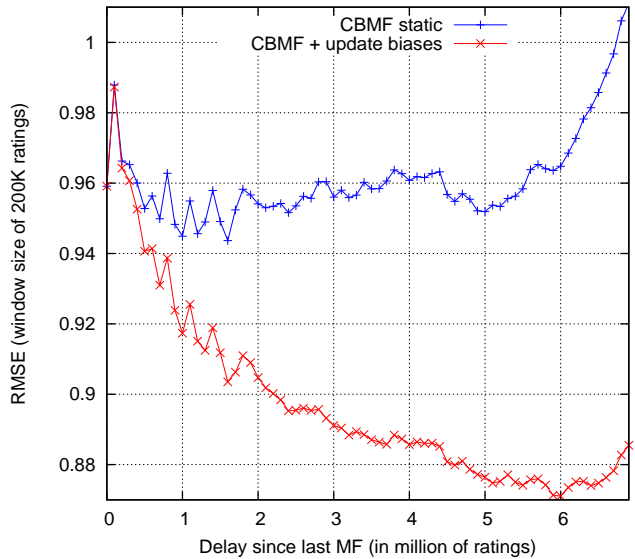}
          \caption{\label{fig:integration} Quality of online integration for increasing delay}
  \end{center}
\end{figure}

The average time to integrate one rating is 0.4 millisecond. It is
fast and adds few overhead on the online recommendation task. As a comparison,
it allows for integrating
more than 216 million ratings per day, which is 27 times bigger than
the Netflix need reported in~\cite{netflix-forum::beyond_the_5_stars_part1}.  In Figure
\ref{fig:integration}, we report the new evolution of CBMF prediction
quality when we integrate the incoming ratings taken from the test
set.

We first analyze the CBMF errors in Figure \ref{fig:integration}, and compare it with the  static (offline) case, to figure out the importance of online integration. 
The benefit of online integration is up to 13.97\% for the largest delay (close to 7M missing ratings), which is a significant improvement for recommendation purpose. That makes our solution quite robust. 


\subsection{Quality vs. Performance tradeoff for online integration}
\begin{figure}
  \begin{center}
           \includegraphics[scale=0.5]{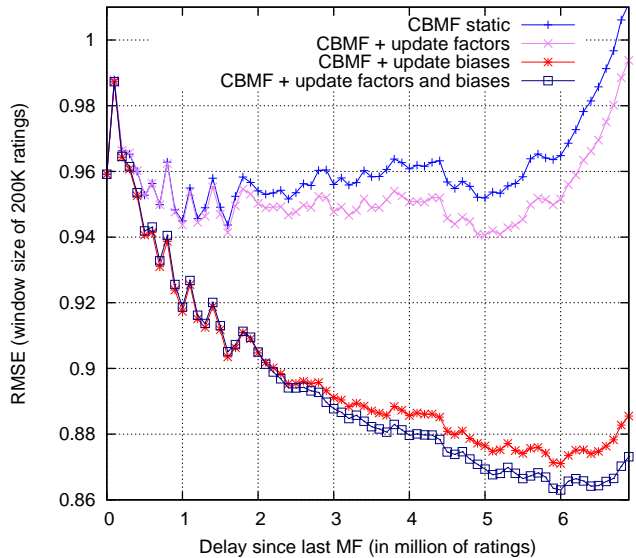}
          \caption{\label{fig:tradeoff} Quality vs. Performance tradeoff}
  \end{center}
\end{figure}

We conducted further experimentations to validate our choice about what part of the model is worth being updated during the online integration phase. We investigated three possible methods to integrate a new rating: (i) update the user factors only, (ii) update the user local biases only, and (iii) update both the user factors and local biases. Naturally, processing more updates comes at a cost. We wondered if the computation time spent in more integration would be eventually amortized by the benefit of postponing the next model re-computation. Figure \ref{fig:tradeoff} shows the quality improvements brought by these three methods of integration.

We reported, on Table \ref{tab:tradeoff} the update time and the respective mean quality gain (in terms of RMSE) for each of the three above mentioned integration methods.
We deduced that integrating both the local biases and the factors bring a relative benefit of 7\% compared to integrating the local biases only. On the other hand, it adds up to 151\% relative overhead on the computation cost. Given a tolerated RMSE value, and the absolute values of the matrix factorization cost and the integration cost, we were able to decide which method yields the minimum overall cost. Table \ref{tab:tradeoff} shows that the local biases-only update method provided the optimal performance (best balance between quality improvement and update cost).
\begin{table}
  \begin{center}
	  \caption{\label{tab:tradeoff} Quality vs. Performance tradeoff}
		\begin{tabular}{|l|c|c|}
		 \hline
		  Update& Improvement (\%)& Average update time \\ \hline
		  user factors & 0.84 &	3.11 ms \\ \hline
		  local biases & 7.18 &	1.24 ms \\ \hline
		  both & 7.69 & 3.75 ms\\ \hline
		\end{tabular}
	\end{center}
\end{table}

\subsection{Benefit of refactorization}
\begin{figure}
  \begin{center}
          \includegraphics[scale=0.5]{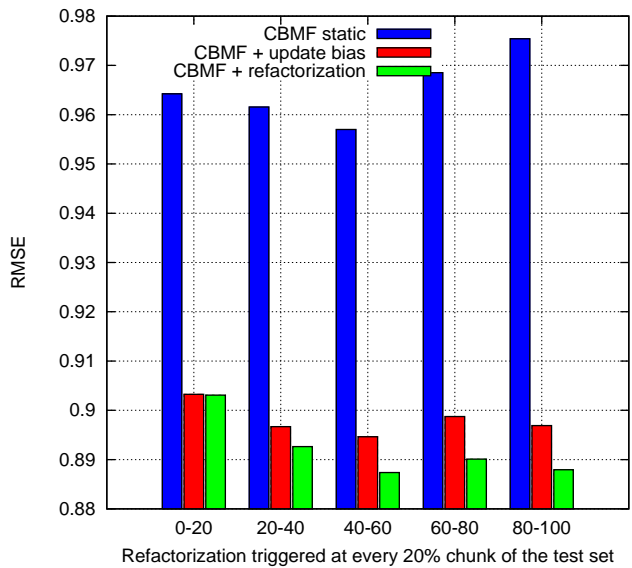}
          \caption{\label{fig:refact} Refactorization benefit}
  \end{center}
\end{figure}

The objective of this experiment is to quantify the benefit of recomputing the CBMF model.
Intuitively, one wish to recompute the model when its quality moves away beyond the expected quality level. On the other hand, in order to save computation resources, we do not wish to recompute the model unless necessary.
With this in mind, we setup an experiment which consists of five successive factorizations. We begin with the same test set and training set as in the previous experiment: the 10\% most recent ratings are in the test set, the remaining 90\% are in the training set. We generate five models resulting from five successive factorizations, scattered in time as described in the following. Let $M_0$ denote the initial model resulting from the training set factorization. Then, we sequentially scan the test set, integrating the incoming ratings into $M_0$, on the fly, until we reach 20\% of the test set. At this point, we trigger the re-factorization and generate a new model, denoted $M_1$, which replaces $M_0$ to become the current model. Then, we repeat the sequence "scan next 20\%, refactorize and replace model" until we reach the end of the test set. We end up generating $M_2$, $M_3$, and $M_4$ which integrate respectively 40 \%, 60\%, and 80\% of the test set in addition to the initial training set.
We report the resulting RMSE, on Figure \ref{fig:refact}, while iterating over the test set and using the most current model, namely $M_0$ to $M_4$, for prediction. We compute each RMSE value based on all the ratings that occur between the current factorization and the next one.

We first globally observe that re-factorization outperforms \emph{CBMF online} at any point in time. Indeed, whatever is the amount of information, a globally optimized model (i.e., factorization) is more accurate than a locally adjusted model (i.e., bias update).
Second, we measure that re-factorization slightly improves \emph{CBMF online} up to 1\% for $M_4$. This is mainly because CBMF online performs quite well all along the run. Hopefully, this offers enough time to recompute the model. In our case, $M_1$ took 8 hours to compute which is the equivalent time to receive 1.33 million ratings (according to the Netflix rate~\cite{netflix-forum::beyond_the_5_stars_part1}). We observe that CBMF online yields low RMSE during a longer time than the time required for re-factorization. This make our solution practical.
Furthermore, a longer run, could serve to measure the maximum "validity time" of the CBMF online. In turn, this would allow to deduce the optimal date to trigger the refactorization, while keeping the RMSE bounded.

\section{Related work}
\label{Related_work}
The main contributions of our work are (1) the use of multi-biases in the matrix factorization process and (2) the integration of the incoming ratings by a quick adjustment of their biases. The problem of the integration of the incoming ratings was well investigated in the literature, while the multi-biases approaches where less studied. Especially, previous work on MF-based recommender systems did not consider biases for the integration of incoming ratings, and focuses on the techniques of factorization~\cite{4809245, sarwar02incrementalsvd, DBLP:conf/recsys/RendleS08, Cao:2007:DTL:1625275.1625708}.

In~\cite{4809245} and~\cite{sarwar02incrementalsvd}, the authors deal with "new user/item" problem, which aims at integrating newly registered users and items (and their ratings).
Even though this problem deals with the integration of new rating, its special nature requires specific solutions. In our approach, we only deal with the new ratings of known users and items.

Rendle et al. focus on users (and items) which have small rating profiles~\cite{DBLP:conf/recsys/RendleS08}. They present an approximation method that updates the matrices of an existing model (previously generated by MF). The proposed \emph{UserUpdate} and \emph{ItemUpdate} algorithms retrain the factor vector for the concerned user, or item, and keep all the other entries in the matrix unchanged. The time complexity of this method is $O(\left|V(u,.)\right|.k.t)$, where $k$ is the given number of factors and $t$ the number of iterations. The whole factor vector of the user is retrained (i.e. his rating profile for all the items), which makes their solution more time consuming than ours ($O(\left|V(u, c(i))\right|.t)$, see Section \ref{Formal_evaluation}). They also not consider user biases, which might be very important for the accuracy of the predictions.

Agarwal et al. propose in~\cite{Agarwal:2010:FOL:1835804.1835894} a fast online bilinear factor model (called FOBFM). It uses an offline analysis of item/user features to initialize the online models. Moreover, it computes linear projections that reduces the dimensionality and, in turn, allows to learn fast both user and item factors in an online fashion. Their offline analysis uses a large amount of historical data (e.g., keywords, categories, browsing behavior) and their model needs to online learn both user and item factors in order to integrate the new ratings. So, their technique is much more costly than ours. Furthermore, our approach works even in applications where no item/user features are available which is not proven in the experimentations of the FOBFM model.

Cao et al.~\cite{Cao:2007:DTL:1625275.1625708} point the problem of data dynamicity in latent factors detection approaches. They propose an \emph{online} nonnegative matrix factorization (ONMF) algorithm that detects latent factors and tracks their evolution when the data
evolves. Let us remind that a nonnegative matrix factorization is a factorization where all the factors in both matrices $P$ and $Q$ are positive. They base their solution on the \textit{Full-Rang Decomposition Theorem}, which states that: for two full rank decompositions $P_{1}.Q_{1}$ and $P_{2}.Q_{2}$ of a matrix $R$, there exists one invertible matrix $X$ satisfying $P_{1} = X.P_{2}$ and $Q_{1} = X^{-1}.Q_{2}$. They use this relation to integrate the new ratings. Although the process seems to be relatively fast, its computation time is greater than ours. This is due to the fact that their technique updates the whole profiles of all the users where our solution limits the computations to the bias of the concerned user.

As said above, using multi-biases into matrix factorization models is not yet the subject of a lot of attention. So far, we only know a few number of works close to ours~\cite{Koren:2010:CFT:1721654.1721677,Koenigstein:2011:YMR:2043932.2043964}. In~\cite{Koren:2010:CFT:1721654.1721677} the author models the drift of user behaviours and item popularity. He incorporates temporal dynamics in the biases of both users and items. Thus, he monitors session-based biases where sessions represent successive time periods. His predictions have better accuracy than the static models. Session biases are combined with the global bias for each item and each user. The focus was different in our work. We did not model the temporal dynamics, instead we opted for the refinement of user biases.
The work described in~\cite{Koenigstein:2011:YMR:2043932.2043964} considers the type of the items in addition to the users and items temporal dynamics. 
For instance, in the music domain, an item type might be \textit{artist}, \textit{album}, \textit{track} or \textit{musical genre}. Thus, sessions are considered to build the user biases and temporal dynamics and item types for the item biases. This approach introduces a type-based grouping of the items which can be considered similar to our items clustering. However, in our approach, the groups of items are not determined according to their type, but according to their ratings similarity.

Our experimentations exposed the need of taking into account the incoming ratings as early as possible in order to keep recommendation quality at a good level. Of course, it is obvious that using parallel implementations leads to better computation time, as shown in~\cite{Gemulla:2011:LMF:2020408.2020426,bickson-blogspot::large_scale_matrix_factorization}. As a consequence, the model can be recomputed more frequently.
However, the need of online integration remains necessary for large scale applications with billions of ratings and many millions of incoming ratings each day (Netflix has more than 5 billion user ratings and receives daily 4 milion new ratings from 23 million subscribers~\cite{netflix-forum::beyond_the_5_stars_part1}). For these applications, a tradeoff between recomputation (with a high cost) and online integration (without a significant lost of quality) is probably the best solution.
\section{Conclusion}
\label{Conclusion}
We tackled the collaborative filtering problem of accurately recommending items to users, when incoming ratings are continuously produced and when the only available information is several millions of user/item ratings.
Through years of experimentation campaigns, the recommendation systems community has demonstrated that the model-based solutions achieve the best quality, however such solutions suffer from a major drawback: they are offline. They take as input a snapshot of the ratings at the time the model computation starts. They simply ignore the more recent ratings, skipping possibly meaningful information for better recommendation.

Our challenging goal was then to find a way to enable the integration of the incoming ratings for a well-know model-based recommendation solution requiring heavy computation with billions of ratings~\cite{netflix-forum::beyond_the_5_stars_part1}.
To this end, we refined the matrix-factorization model that features very good offline quality, by introducing personalized biases that capture the user subjectivity for different groups of items. Items being groupe basing on their ratings.

We proposed a detailed algorithm to update the fine grained (i.e. per item cluster) user biases, which is fast enough to integrate the incoming ratings as soon as they are produced.
We implemented the algorithms and performed extensive experiments on two real large datasets containing respectively 10M and 100M ratings, in order to validate both quality and performance of our cluster-based matrix factorization (CBMF) approach. We compared our solution with two state-of-the-art matrix factorization solutions that support 0 and 1 bias respectively. Qualitative results place our solution better to its competitors in the offline case.
Our solution demonstrates an improvement of accuracy up to 13.97\% (relatively to the offline case) for highly dynamic scenario
where millions of incoming ratings are injected into the model. Moreover, performance results expose fast integration of the incoming ratings; which makes our solution viable for online recommendation systems that need to scale up to a higher throughput of incoming ratings.

%

\end{document}